\documentclass[11pt,a4paper]{article}
\usepackage[hyperref]{emnlp2020}
\usepackage{times}
\usepackage{latexsym}

\usepackage{microtype}

\usepackage{graphicx}
\usepackage{amsfonts}
\usepackage{amsmath}
\usepackage{bm}
\usepackage{amssymb}
\usepackage{pifont}
\usepackage{nicefrac}
\usepackage{multirow}
\usepackage{subcaption}
\usepackage[font=small]{caption}
\usepackage{soul}
\usepackage{xcolor}
\usepackage{booktabs}
\usepackage{algorithm}
\usepackage{algpseudocode}
\usepackage{mathtools}
\usepackage{comment}
\usepackage{textcomp}
\usepackage[most]{tcolorbox}

\usepackage{tablefootnote}

\definecolor{bblue}{HTML}{4F81BD}
\definecolor{rred}{HTML}{c4260b}
\definecolor{ggreen}{HTML}{098c1f}
\definecolor{ppurple}{HTML}{9F4C7C}
\definecolor{oorange}{HTML}{F79646}

\algnewcommand\algorithmicinput{\textbf{Input:}}
\algnewcommand\Input{\item[\algorithmicinput]}
\algnewcommand\algorithmicoutput{\textbf{Output:}}
\algnewcommand\Output{\item[\algorithmicoutput]}

\algnewcommand\algorithmicempty{~}
\algnewcommand\Empty{\item[\algorithmicempty]}

\usepackage{enumitem}
\setenumerate[1]{itemsep=0.30pt,partopsep=0pt,parsep=\parskip,topsep=5pt}
\setitemize[1]{itemsep=0.30pt,partopsep=00pt,parsep=\parskip,topsep=5pt}
\setdescription{itemsep=0.30pt,partopsep=0pt,parsep=\parskip,topsep=5pt}

\usepackage{todonotes}

\aclfinalcopy 

\newcommand\afs{AFS}
\newcommand\afst{AFS$^t$}

\newcommand\afstf{AFS$^{t,f}$}
\newcommand\lzerodrop{$\mathcal{L}_0$\textsc{Drop}}

\title{Adaptive Feature Selection for End-to-End Speech Translation}

\author{Biao Zhang$^1$ \quad Ivan Titov$^{1,2}$ \quad Barry Haddow$^1$ \quad Rico Sennrich$^{3,1}$ \bigskip\\
  $^1$School of Informatics, University of Edinburgh \\
  $^2$ILLC, University of Amsterdam \\
  $^3$Department of Computational Linguistics, University of Zurich \\
  \resizebox{\textwidth}{!}{\texttt{B.Zhang@ed.ac.uk, \{ititov,bhaddow\}@inf.ed.ac.uk, sennrich@cl.uzh.ch}}
  }

\date{}

\begin{document}
\maketitle
\begin{abstract}

Information in speech signals is not evenly distributed, making it an additional challenge for end-to-end (E2E) speech translation (ST) to learn to focus on informative features.
In this paper, we propose adaptive feature selection (\afs{}) for encoder-decoder based E2E ST.
We first pre-train an ASR encoder and apply \afs{} to dynamically estimate the importance of each encoded speech feature to ASR.
A ST encoder, stacked on top of the ASR encoder, then receives the filtered features from the (frozen) ASR encoder.
We take \lzerodrop{}~\cite{zhang2020sparsifying} as the backbone for \afs, and adapt it to sparsify speech features with respect to both temporal and feature dimensions. Results on LibriSpeech En-Fr and MuST-C benchmarks show that \afs{} facilitates learning of ST by pruning out $\sim$84\% temporal features, yielding an average translation gain of $\sim$1.3--1.6 BLEU and a decoding speedup of $\sim$1.4$\times$. In particular, \afs{} reduces the performance gap compared to the cascade baseline, and outperforms it on LibriSpeech En-Fr with a BLEU score of 18.56 (without data augmentation).\footnote{We release our source code at \url{https://github.com/bzhangGo/zero}.}

\end{abstract}

\section{Introduction}

End-to-end (E2E) speech translation (ST), a paradigm that directly maps audio to a foreign text, has been  gaining popularity recently~\cite{duong-etal-2016-attentional,berard2016listen,bansal2018low,di2019adapting,wang2019bridging}. Based on the attentional encoder-decoder framework~\cite{DBLP:journals/corr/BahdanauCB14}, it optimizes model parameters under direct translation supervision. This end-to-end paradigm avoids the problem of error propagation that is inherent in cascade models where an automatic speech recognition (ASR) model and a machine translation (MT) model are 
chained together. Nonetheless, previous work still reports that E2E ST delivers inferior performance compared to cascade methods~\cite{niehues_j_2019_3532822}.

\begin{figure}[t]
  \centering
  \includegraphics[scale=0.385]{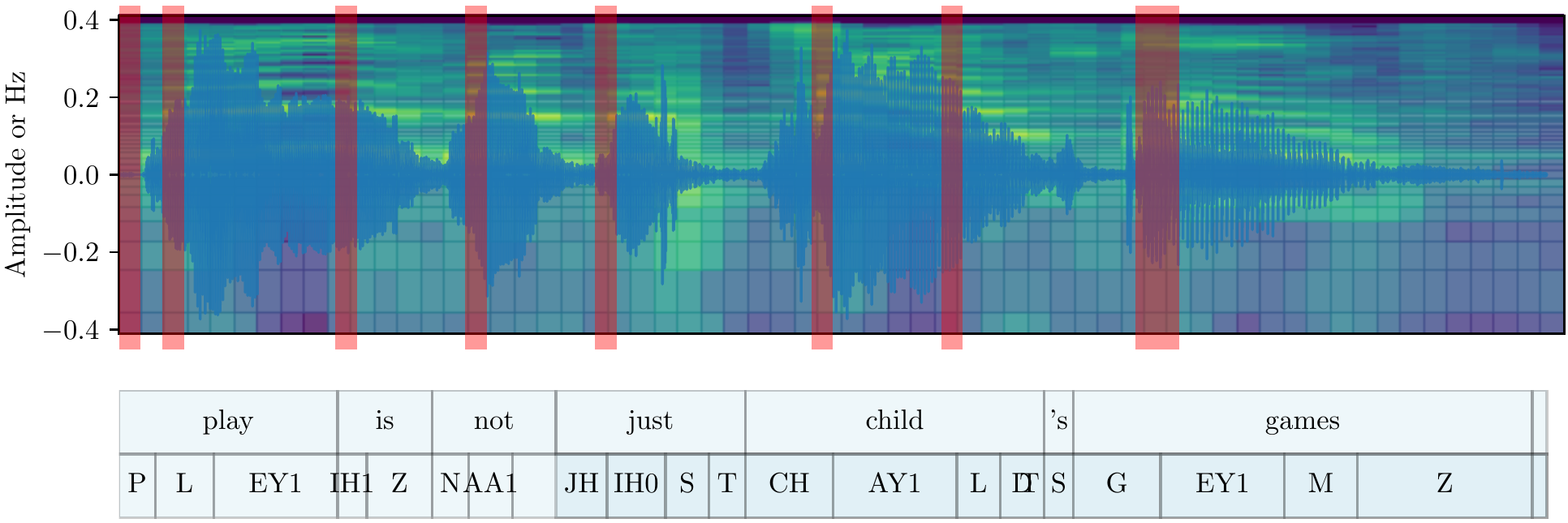}
  \caption{\label{fig:illustration} Example illustrating our motivation. We plot the amplitude and frequency spectrum of an audio segment (top), paired with its time-aligned words and phonemes (bottom). Information inside an audio stream is not uniformly distributed. We propose to dynamically capture speech features corresponding to informative signals (red rectangles) to improve ST.}
\end{figure}

We study one reason for the difficulty of training E2E ST models, namely the uneven spread of information in the speech signal, as visualized in Figure \ref{fig:illustration}, and the consequent difficulty of extracting informative features.
Features corresponding to uninformative signals, such as pauses or noise, increase the input length and bring in unmanageable noise for ST.
This increases the difficulty of learning~\cite{Zhang2019trainable,na2019adaptive} and reduces translation performance.

\begin{figure}[t]
  \centering
  \includegraphics[scale=0.45]{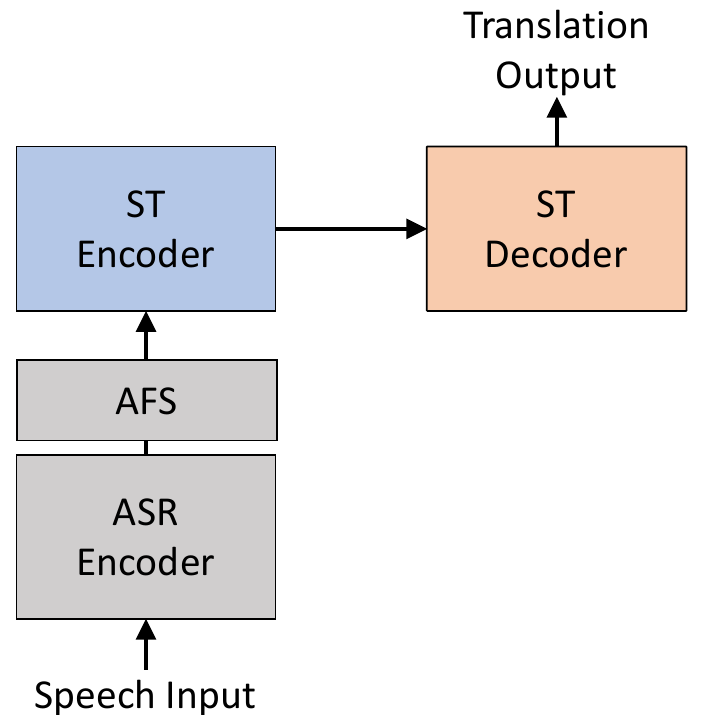}
  \caption{\label{fig:overview} Overview of our E2E ST model. \afs{} is inserted between the ST encoder (blue) and a pretrained ASR encoder (gray) to filter speech features for translation. We pretrain \afs{} jointly with ASR and freeze it during ST training.}
\end{figure}

In this paper, we propose adaptive feature selection (\afs{}) for ST to explicitly eliminate uninformative features. Figure \ref{fig:overview} shows the overall architecture. We employ a pretrained ASR encoder to induce contextual speech features, followed by an ST encoder bridging the gap between speech and translation modalities.
\afs{} is inserted in-between them to select a subset of features for ST encoding (see red rectangles in Figure \ref{fig:illustration}). To ensure that the selected features are well-aligned to transcriptions, we pretrain \afs{} on ASR. \afs{} estimates the informativeness of each feature through a parameterized gate, and encourages the dropping of features (pushing the gate to $0$) that contribute little to ASR. An underlying assumption is that features irrelevant for ASR are also unimportant for ST.

We base \afs{} on \lzerodrop{}~\cite{zhang2020sparsifying}, a sparsity-inducing method for encoder-decoder models, and extend it to sparsify speech features. The acoustic input of speech signals involves two dimensions: \textit{temporal} and \textit{feature}, where the latter one describes the spectrum extracted from time frames.  
Accordingly, we adapt \lzerodrop{} to sparsify encoder states along temporal and feature dimensions but using different gating networks.
In contrast to \cite{zhang2020sparsifying}, who focus on efficiency and report a trade-off between sparsity and quality for MT and summarization, we find that sparsity also improves translation quality for ST. 

We conduct extensive experiments with Transformer~\cite{NIPS2017_7181_attention} on LibriSpeech En-Fr and MuST-C speech translation tasks, covering 8 different language pairs. Results show that \afs{} only retains about 16\% of temporal speech features, revealing heavy redundancy in speech encodings and yielding a decoding speedup of $\sim$1.4$\times$. \afs{} eases model convergence, and improves the translation quality by $\sim$1.3--1.6 BLEU, surpassing several strong baselines. 
Specifically, without data augmentation, \afs{} narrows the performance gap against the cascade approach, and outperforms it on LibriSpeech En-Fr by 0.29 BLEU, reaching 18.56.
We compare against fixed-rate feature selection and a simple CNN, confirming that our adaptive feature selection offers better translation quality. 

Our work demonstrates that E2E ST suffers from redundant speech features, with sparsification bringing significant performance improvements. The E2E ST task offers new opportunities for follow-up research in sparse models to deliver performance gains, apart from enhancing efficiency and/or interpretability.

\section{Background: \lzerodrop{}}

\lzerodrop{} provides a selective mechanism for encoder-decoder models which encourages removing uninformative encoder outputs via a sparsity-inducing  objective~\cite{zhang2020sparsifying}. Given a source sequence $X=\{x_1, x_2, \ldots, x_n\}$, \lzerodrop{} assigns each encoded source state $\mathbf{x}_i \in \mathbb{R}^{d}$ with a scalar gate $g_i \in [0, 1]$ as follows:
\begin{align}
    \text{\lzerodrop{}}(\mathbf{x}_i) = g_i \mathbf{x}_i, \label{eq:l0drop_formula} \\
    \text{with~}~ g_i \sim \text{HardConcrete}(\alpha_i, \beta, \epsilon),
\end{align}
where $\alpha_i, \beta, \epsilon$ are hyperparameters of the hard concrete distribution (HardConcrete)~\cite{louizos2017learning}.

Note that the hyperparameter $\alpha_i$ is crucial to HardConcrete as
it directly governs its shape. We associate $\alpha_i$ with $\mathbf{x}_i$ through a gating network: 
\begin{equation}
    \log \alpha_i = \mathbf{x}_i^T \cdot \mathbf{w}, \label{eq:l0_gate_network}
\end{equation}
Thus, \lzerodrop{} can schedule HardConcrete via $\alpha_i$ to put more probability mass at either $0$ (i.e $g_i \rightarrow 0$) or $1$ (i.e. $g_i \rightarrow 1$). $\mathbf{w} \in \mathbb{R}^d$ is a trainable parameter. Intuitively, \lzerodrop{} controls the openness of gate $g_i$ via $\alpha_i$ so as to determine whether to remove ($g_i=0$) or retain ($g_i=1$) the state $\mathbf{x}_i$.

\lzerodrop{} enforces sparsity by pushing the probability mass of HardConcrete towards $0$, according to the following penalty term:
\begin{equation}
    \mathcal{L}_0(X) = \sum_{i=1}^{n}{1 - p(g_i = 0 | \alpha_i, \beta, \epsilon )}. \label{eq:l0drop_sparsity_penalty}
\end{equation}
By sampling $g_i$ with reparameterization~\cite{kingma2013auto}, \lzerodrop{} is fully differentiable and optimized with an upper bound on the objective: $\mathcal{L}_{\textsc{mle}} + \lambda \mathcal{L}_0(X)$, 
where $\lambda$ is a hyperparameter affecting the degree of sparsity -- a larger $\lambda$ enforces more gates near 0 -- and $\mathcal{L}_{\textsc{mle}}$ denotes the maximum likelihood loss. An estimation of the expected value of $g_i$ is used during inference. 
\citet{zhang2020sparsifying} applied \lzerodrop{} to prune encoder outputs for MT and summarization tasks; we adapt it to E2E ST. 
Sparse stochastic  gates and 
$\mathcal{L}_0$ relaxations were also by \citet{bastings-etal-2019-interpretable} to construct interpretable classifiers, i.e. models that can reveal
which tokens they rely on when making a prediction.

\section{Adaptive Feature Selection}

One difficulty with applying encoder-decoder models to E2E ST is deciding how to encode speech signals. In contrast to text where word boundaries can be easily identified, the spectrum features of speech are continuous, varying remarkably across different speakers for the same transcript. 
In addition, redundant information, like pauses in-between neighbouring words, can be of arbitrary duration at any position as shown in Figure \ref{fig:illustration}, while  contributing little to translation. This increases the burden and occupies the capacity of ST encoder, leading to inferior performance~\cite{duong-etal-2016-attentional,berard2016listen}.
Rather than developing complex encoder architectures, we resort to feature selection to explicitly clear out those uninformative speech features. 

Figure \ref{fig:overview} gives an overview of our model. We use a pretrained and frozen ASR encoder to extract contextual speech features, and collect the informative ones from them via \afs{} before transmission to the ST encoder. \afs{} drops pauses, noise and other uninformative features and retains features that are relevant for ASR. 
We speculate that these retained features are also the most relevant for ST, and that the sparser representation simplifies the learning problem for ST, for example the learning of attention strength between encoder states and target language (sub)words.
Given a training tuple (audio, source transcription, translation), denoted as $(X, Y, Z)$ respectively,\footnote{Note that our model only requires pair-wise training corpora, $(X, Y)$ for ASR, and $(X, Z)$ for ST.} we outline the overall framework below, including three steps:

\begin{tcolorbox}[enhanced,attach boxed title to top center={yshift=-3mm,yshifttext=-1mm}, title=E2E ST with AFS, colback=white, colframe=black, coltitle=black, colbacktitle=white, toprule=0.5pt, bottomrule=0.5pt,leftrule=0.5pt,rightrule=0.5pt]
    \fontsize{9.5}{11}\selectfont
    \begin{enumerate}[leftmargin=*]
        \item Train ASR model with the following objective and model architecture until convergence:
            \begin{align}
                \mathcal{L}^{\text{ASR}} & = \eta \mathcal{L}_{\textsc{mle}}(Y|X) + \gamma \mathcal{L}_{\textsc{ctc}}(Y|X), \label{eq:asr_objective}\\
               \mathcal{M}^{\text{ASR}} & = D^{\text{ASR}}\left(Y, E^{\text{ASR}}\left(X\right)\right). \label{eq:asr_model}
            \end{align}
        \item Finetune ASR model with AFS for $m$ steps:
            \begin{align}
                \mathcal{L}^{\text{AFS}} & =\mathcal{L}_{\textsc{mle}}(Y|X) + \lambda \mathcal{L}_0(X), \label{eq:afs_objective} \\
               \mathcal{M}^{\text{AFS}} & = D^{\text{ASR}}\left(Y, F\left(E^{\text{ASR}}\left(X\right)\right)\right). \label{eq:afs_model}
            \end{align}
        \item Train ST model with pretrained and frozen ASR and AFS submodules until convergence:
            \begin{align}
                \mathcal{L}^{\text{ST}} & = \mathcal{L}_{\textsc{mle}}(Z|X), \label{eq:st_objective} \\
                \mathcal{M}^{\text{ST}} & = D^{\text{ST}}\left(Z, E^{\text{ST}}\left(\overline{F}
                \overline{E}^{\text{ASR}}\left(X\right)\right)\right). \label{eq:st_model}
            \end{align}
    \end{enumerate}
\end{tcolorbox}

\noindent We handle both ASR and ST as sequence-to-sequence problem with encoder-decoder models. We use $E^{*}(\cdot)$ and $D^{*}(\cdot, \cdot)$ to denote the corresponding encoder and decoder respectively. $F(\cdot)$ denotes the \afs{} approach, and $\overline{FE}$ means freezing the ASR encoder and the \afs{} module during training. Note that our framework puts no constraint on the architecture of the encoder and decoder in any task, although we adopt the multi-head dot-product attention network~\cite{NIPS2017_7181_attention} for our experiments.

\paragraph{ASR Pretraining} The ASR model $\mathcal{M}^{\text{ASR}}$ (Eq. \ref{eq:asr_model}) directly maps an audio input to its transcription. To improve speech encoding, we apply logarithmic penalty on attention to enforce short-range dependency~\cite{di2019adapting} and use trainable positional embedding with a maximum length of 2048. Apart from $\mathcal{L}_{\textsc{mle}}$, we augment the training objective with the connectionist temporal classification~\cite[CTC]{Graves06connectionisttemporal} loss $\mathcal{L}_{\textsc{ctc}}$ as in Eq. \ref{eq:asr_objective}. Note $\eta = 1 - \gamma$. The CTC loss is applied to the encoder outputs, guiding them to align with their corresponding transcription (sub)words and improving the encoder's robustness~\cite{karita2019transformerasr}. Following previous work~\cite{karita2019transformerasr,wang2020curriculum}, we set $\gamma$ to $0.3$.

\paragraph{\afs{} Finetuning} This stage aims at using \afs{} to dynamically pick out the subset of ASR encoder outputs that are most relevant for ASR performance (see red rectangles in Figure \ref{fig:illustration}).
We follow \citet{zhang2020sparsifying} and place \afs{} in-between ASR encoder and decoder during finetuning (see $F(\cdot)$ in $\mathcal{M}^{\text{AFS}}$, Eq. \ref{eq:afs_model}). We exclude the CTC loss in the training objective (Eq. \ref{eq:afs_objective}) to relax the alignment constraint and increase the flexibility of feature adaptation. 
We use \lzerodrop{} for \afs{} in two ways. 

\textbf{\afst{}} The direct application of \lzerodrop{} on ASR encoder results in \afst{}, sparsifying encodings along the temporal dimension $\{\mathbf{x}_i\}_{i=1}^n$:
\begin{equation}\label{eq:afs_t}
    \begin{split}
        F^t(\mathbf{x}_i) & = \text{\afst{}}(\mathbf{x}_i) = g^t_{i} \mathbf{x}_i, \\
        \text{with~}~ \log \alpha^t_{i} & = \mathbf{x}_i^T \cdot \mathbf{w}^t, \\
        g^t_{i} & \sim \text{HardConcrete}(\alpha^t_{i}, \beta, \epsilon),
    \end{split}
\end{equation}
where $\alpha^t_{i}$ is a positive scalar powered by a simple linear gating layer, and $\mathbf{w}^t \in \mathbb{R}^{d}$ is a trainable parameter of dimension $d$. $\mathbf{g}^t$ is the temporal gate. The sparsity penalty of \afst{} follows Eq. \ref{eq:l0drop_sparsity_penalty}:
\begin{equation}\label{eq:afs_t_l0}
    \mathcal{L}_{0}^t (X) = \sum_{i=1}^{n}{1 - p(g^t_{i} = 0 | \alpha^t_{i}, \beta, \epsilon )}.
\end{equation}

\textbf{\afstf{}} In contrast to text processing, speech processing often extracts spectrum from overlapping time frames to form the acoustic input, similar to the word embedding. As each encoded speech feature contains temporal information, it is reasonable to extend \afst{} to \afstf{}, including sparsification along the feature dimension $\{\mathbf{x}_{i,j}\}_{j=1}^d$:
\begin{equation}\label{eq:afs_tf}
    \begin{split}
        F^{t,f}(\mathbf{x}_i) & = \text{\afstf{}}(\mathbf{x}_i) = g^t_{i} \mathbf{x}_i \odot \mathbf{g}^{f}, \\
        \text{with~}~ \log {\bm \alpha}^{f} & = \mathbf{w}^f, \\
        {g}^f_{j} & \sim \text{HardConcrete}({\alpha}^f_{j}, \beta, \epsilon),
    \end{split}
\end{equation}
where ${\bm \alpha}^{f} \in \mathbb{R}^{d}$ estimates the weights of each feature, dominated by an input-independent gating model with trainable parameter $\mathbf{w}^f \in \mathbb{R}^d$.\footnote{Other candidate gating models, like linear mapping upon mean-pooled encoder outputs, delivered worse performance in our preliminary experiments.} $\mathbf{g}^f$ is the feature gate. Note that ${\bm \alpha}^{f}$ is shared for all time steps. $\odot$ denotes element-wise multiplication. \afstf{} reuses $g^t_{i}$-relevant submodules in Eq. \ref{eq:afs_t}, and extends the sparsity penalty $\mathcal{L}^t_{0}$ in Eq. \ref{eq:afs_t_l0} as follows:
\begin{equation}\label{eq:afs_tf_l0}
    \mathcal{L}^{t,f}_{0} (X) = \mathcal{L}^t_{0} + \sum_{j=1}^{d}{1 - p({g}^f_{j} = 0 | {\alpha}^f_{j}, \beta, \epsilon )}.
\end{equation}
We perform the finetuning by replacing ($F, \mathcal{L}_0$) in Eq. (\ref{eq:afs_model}-\ref{eq:afs_objective}) with either \afst{} ($F^t, \mathcal{L}^t_0$) or \afstf{} ($F^{t,f}, \mathcal{L}^{t,f}_0$) for extra $m$ steps. We compare these two variants in our experiments. 

\paragraph{E2E ST Training} We treat the pretrained ASR and \afs{} model as a speech feature extractor, and freeze them during ST training. We gather the speech features emitted by the ASR encoder that correspond to $g^t_i > 0$, and pass them similarly as done with word embeddings to the ST encoder.
We employ sinusoidal positional encoding to distinguish features at different positions. Except for the input to the ST encoder, our E2E ST follows the standard encoder-decoder translation model ($\mathcal{M}^{\text{ST}}$ in Eq. \ref{eq:st_model}) and is optimized with $\mathcal{L}_{\textsc{mle}}$ alone as in Eq. \ref{eq:st_objective}. Intuitively, \afs{} bridges the gap between ASR output and MT input by selecting transcript-aligned speech features.

\section{Experiments}

\paragraph{Datasets and Preprocessing}
We experiment with two benchmarks: the Augmented LibriSpeech dataset (LibriSpeech En-Fr)~\cite{kocabiyikoglu-etal-2018-augmenting} and the multilingual MuST-C dataset (MuST-C)~\cite{di-gangi-etal-2019-must}.
LibriSpeech En-Fr is collected by aligning e-books in French with English utterances of LibriSpeech, further augmented with French translations offered by Google Translate. We use the 100 hours clean training set for training, including 47K utterances to train ASR models and double the size for ST models after concatenation with the Google translations. We report results on the test set (2048 utterances) using models selected on the dev set (1071 utterances).
MuST-C is built from English TED talks, covering 8 translation directions: English to German (De), Spanish (Es), French (Fr), Italian (It), Dutch (Nl), Portuguese (Pt), Romanian (Ro) and Russian (Ru). We train ASR and ST models on the given training set, containing $\sim$452 hours with $\sim$252K utterances on average for each translation pair. We adopt the given dev set for model selection and report results on the common test set, whose size ranges from 2502 (Es) to 2641 (De) utterances.

For all datasets, we extract 40-dimensional log-Mel filterbanks with a step size of 10ms and window size of 25ms as the acoustic features. We expand these features with their first and second-order derivatives, and stabilize them using mean subtraction and variance normalization.
We stack the features corresponding to three consecutive frames without overlapping to the left, resulting in the final 360-dimensional acoustic input. For transcriptions and translations, we tokenize and truecase all the text using Moses scripts \cite{koehn-etal-2007-moses}. We train subword models~\cite{sennrich-etal-2016-neural} on each dataset with a joint vocabulary size of 16K to handle rare words, and share the model for ASR, MT and ST. We train all models without removing punctuation.

\paragraph{Model Settings and Baselines}
We adopt the Transformer architecture~\cite{NIPS2017_7181_attention} for all tasks, including $\mathcal{M}^{\text{ASR}}$ (Eq. \ref{eq:asr_model}), $\mathcal{M}^{\text{AFS}}$ (Eq. \ref{eq:afs_model}) and $\mathcal{M}^{\text{ST}}$ (Eq. \ref{eq:st_model}). The encoder and decoder consist of 6 identical layers, each including a self-attention sublayer, a cross-attention sublayer (decoder alone) and a feedforward sublayer. We employ the base setting for experiments: hidden size $d =512$, attention head 8 and feedforward size 2048. 
We schedule learning rate via Adam ($\beta_1=0.9, \beta_2=0.98$)~\cite{kingma2014adam}, paired with a warmup step of 4K. We apply dropout to attention weights and residual connections with a rate of 0.1 and 0.2 respectively, and also add label smoothing of 0.1 to handle overfitting. We train all models with a maximum step size of 30K and a minibatch size of around 25K target subwords. We average the last 5 checkpoints for evaluation. We use beam search for decoding, and set the beam size and length penalty to 4 and 0.6, respectively.
We set $\epsilon=-0.1$, and $\beta=\nicefrac{2}{3}$ for \afs{} following~\citet{louizos2017learning}, and finetune \afs{} for an additional $m=5\text{K}$ steps.
We evaluate translation quality with tokenized case-sensitive BLEU~\cite{papineni-etal-2002-bleu}, and report WER for ASR performance without punctuation. 

We compare our models with four baselines:
\begin{description}
    \item[\textbf{ST}:] A vanilla Transformer-based E2E ST model of 6 encoder and decoder layers. Logarithmic attention penalty~\cite{di2019adapting} is used to improve the encoder.
    \item[\textbf{ST + ASR-PT}:] We perform the ASR pretraining (ASR-PT) for E2E ST. This is the same model as ours (Figure \ref{fig:overview}) but without \afs{} finetuning. 
    \item[\textbf{Cascade}:] We first transcribe the speech input using an ASR model, and then passes the results on to an MT model. We also use the logarithmic attention penalty~\cite{di2019adapting} for the ASR encoder.
    \item[\textbf{ST + Fixed Rate}:] Instead of dynamically selecting features, we replace \afs{} with subsampling at a fixed rate: we extract the speech encodings after every $k$ positions. 
\end{description}
Besides, we offer another baseline, \textbf{ST + CNN}, for comparison on MuST-C En-De: we replace the fixed-rate subsampling with a one-layer 1D depth-separable convolution, where the output dimension is set to 512, the kernel size over temporal dimension is set to 5 and the stride is set to 6. In this way, the ASR encoder features will be compressed to around 1/6 features, a similar ratio to the fixed-rate subsampling.

\subsection{Results on MuST-C En-De}

\begin{figure}[t]
  \centering
    \subcaptionbox{\label{fig:ana_lambda:freq} Feature Gate Value }[0.494\columnwidth]{
        \centering
        \includegraphics[scale=0.40]{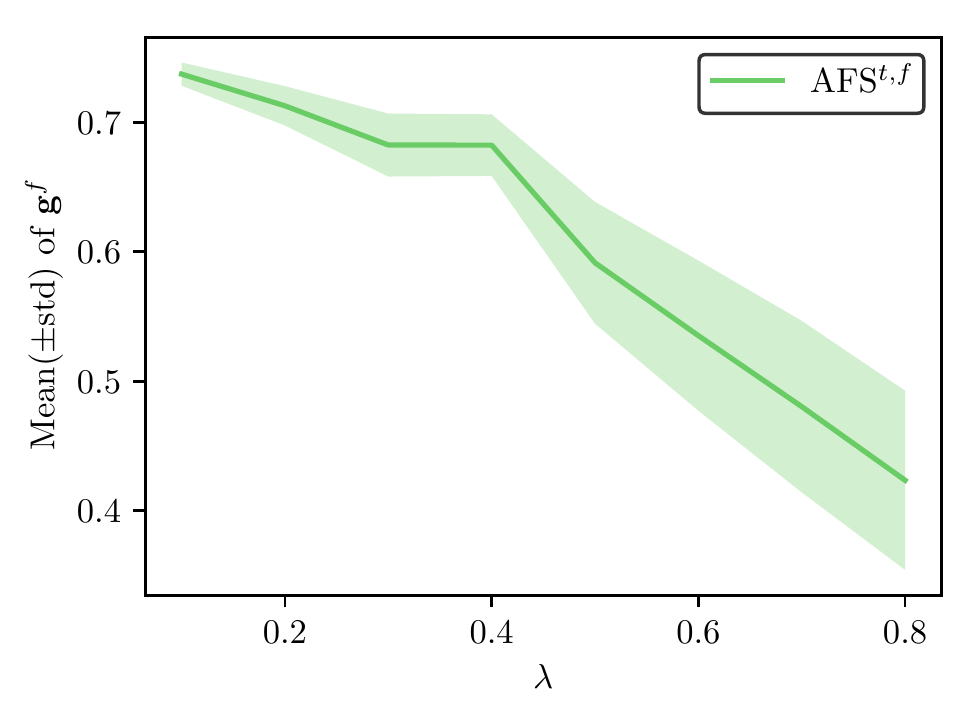}
    }
    \subcaptionbox{\label{fig:ana_lambda:temp} Temporal Sparsity Rate}[0.494\columnwidth]{
        \centering
        \includegraphics[scale=0.40]{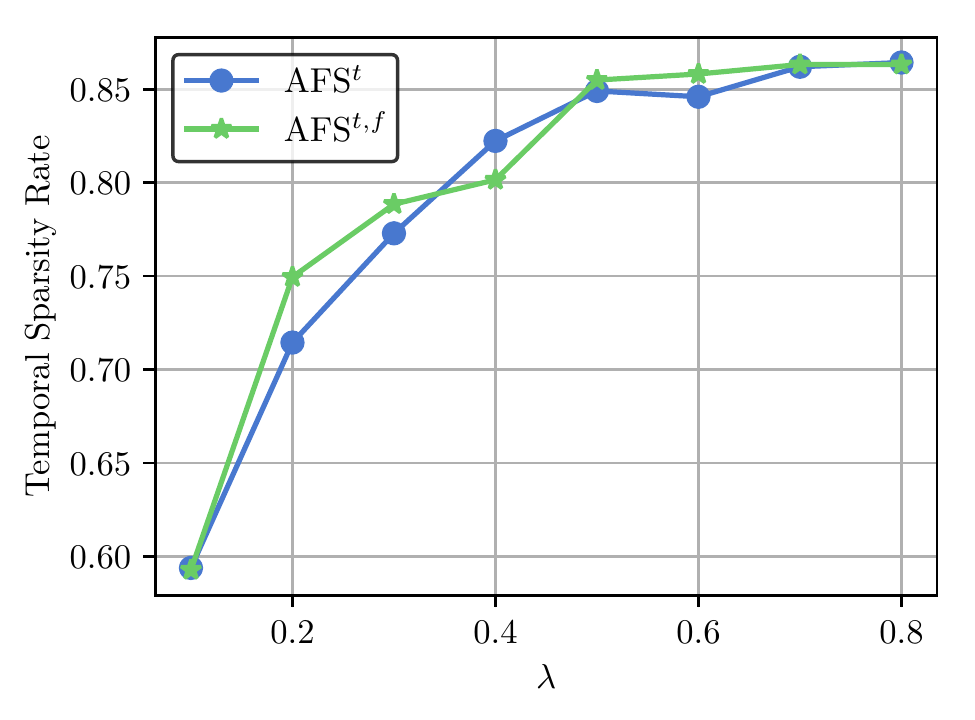}
    }
  \caption{\label{fig:ana_lambda} Feature gate value and temporal sparsity rate as a function of $\lambda$ on MuST-C En-De dev set. Larger $\lambda$ decreases the gate value of $\mathbf{g}^f$ but without dropping any neurons, i.e. feature sparsity rate 0\%. By contrast, speech features are of high redundancy along temporal dimension, easily inducing high sparsity rate of $\sim$85\%.}
\end{figure}

We perform a thorough study on MuST-C En-De. With \afs{}, the first question is its feasibility. We start by analyzing the degree of sparsity in speech features (i.e. sparsity rate) yielded by \afs{}, focusing on the temporal sparsity rate $\nicefrac{\#\{g^t_{i} = 0\}}{n}$ and the feature sparsity rate $\nicefrac{\#\{g^f_{j} = 0\}}{d}$. To obtain different rates, we vary the hyperparameter $\lambda$ in Eq. \ref{eq:afs_objective} in a range of $[0.1, 0.8]$ with a step size 0.1.

Results in Figure \ref{fig:ana_lambda} show that large amounts of encoded speech features ($>59\%$) can be easily pruned out, revealing heavy inner-speech redundancy. Both \afst{} and \afstf{} drop $\sim$60\% temporal features with $\lambda$ of 0.1, and this number increases to $>85\%$ when $\lambda \geq 0.5$ (Figure \ref{fig:ana_lambda:temp}), remarkably surpassing the sparsity rate reported by \citet{zhang2020sparsifying} on text summarization ($71.5\%$). In contrast to rich temporal sparsification, we get a feature sparsity rate of 0, regardless of $\lambda$'s value, although increasing $\lambda$ decreases $\mathbf{g}^f$ (Figure \ref{fig:ana_lambda:freq}). This suggests that selecting neurons from the feature dimension is harder. Rather than filtering neurons, the feature gate $\mathbf{g}^f$ acts more like a weighting mechanism on them. In the rest of the paper, we use \textit{sparsity rate} for the temporal sparsity rate.

\begin{figure}[t]
  \centering
  \small
    \subcaptionbox{\label{fig:ana_quality:wer} ASR}[0.494\columnwidth]{
        \centering
        \includegraphics[scale=0.40]{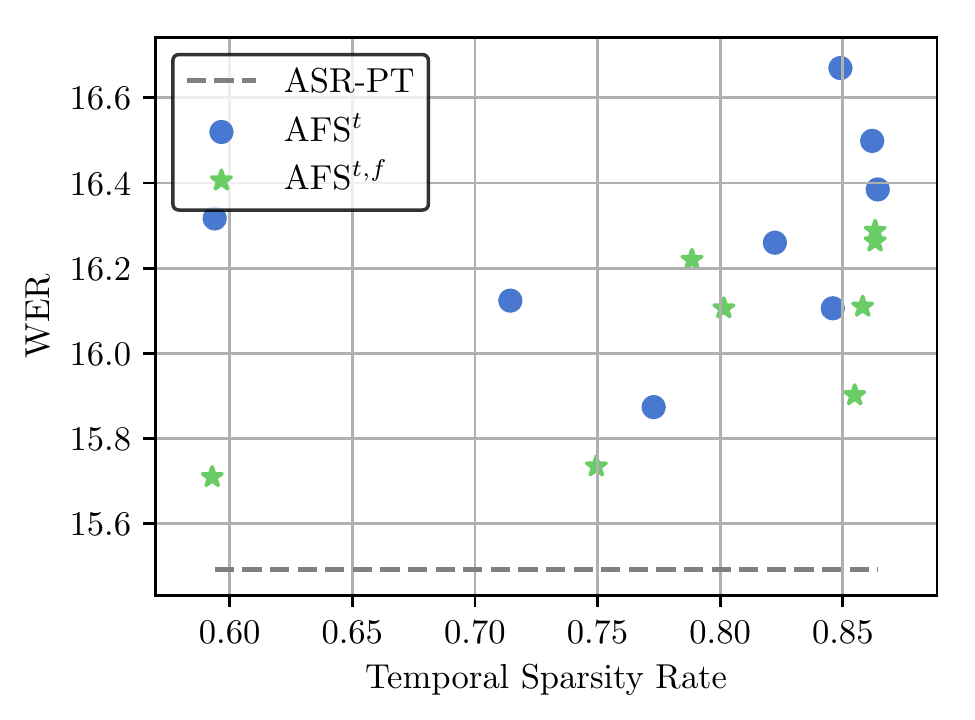}
    }
    \subcaptionbox{\label{fig:ana_quality:bleu} ST}[0.494\columnwidth]{
        \centering
        \includegraphics[scale=0.40]{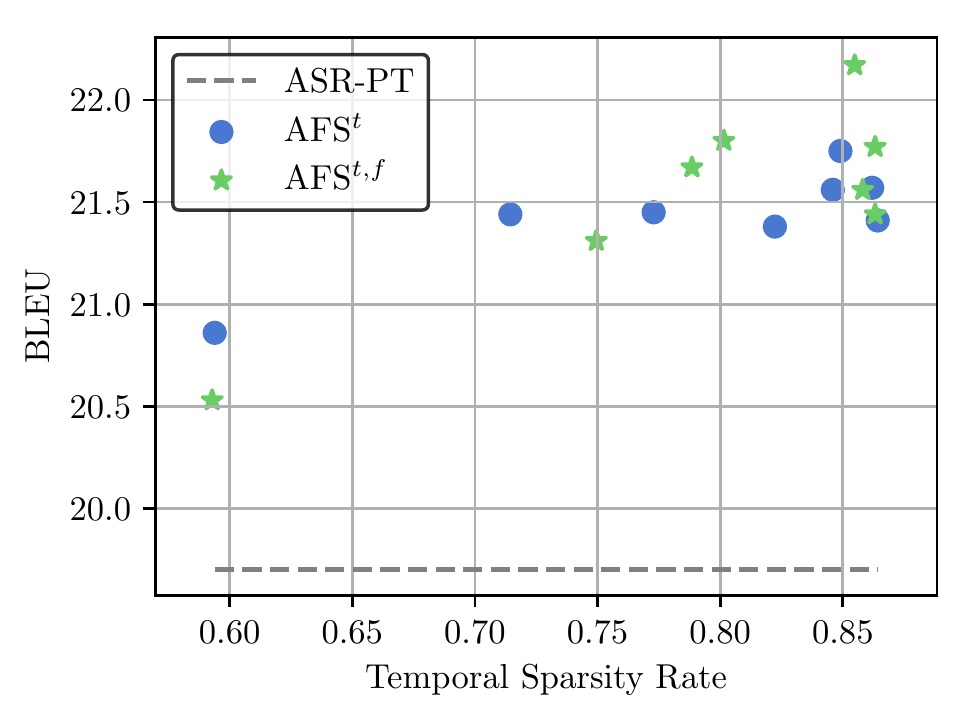}
    }
  \caption{\label{fig:ana_quality} ASR (WER$\downarrow$) and ST (BLEU$\uparrow$) performance as a function of \textit{temporal} sparsity rate on MuST-C En-De dev set. Pruning out $\sim$85\% temporal speech features largely improves translation quality and retains $\sim$95\% ASR accuracy.}
\end{figure}

We continue to explore the impact of varied sparsity rates on the ASR and ST performance. Figure \ref{fig:ana_quality} shows their correlation. We observe that \afs{} slightly degenerates ASR accuracy (Figure \ref{fig:ana_quality:wer}), but still retains $\sim$95\% accuracy on average; \afstf{} often performs better than \afst{} with similar sparsity rate. The fact that only $15\%$ speech features successfully support 95\% ASR
accuracy proves the informativeness of these selected features.
These findings echo with \cite{zhang2020sparsifying}, where they observe a trade-off between sparsity and quality.

However, when \afs{} is applied to ST, we find consistent improvements to translation quality by $>0.8$ BLEU, shown in Figure \ref{fig:ana_quality:bleu}.
Translation quality on the development set peaks at 22.17 BLEU achieved by \afstf{} with a sparsity rate of 85.5\%. We set $\lambda=0.5$ (corresponding to sparsity rate of $\sim$85\%) for all other experiments, since \afst{} and \afstf{} reach their optimal result at this point.

\begin{table}[t]
    \centering
    \small
    \begin{tabular}{lcc}
      \toprule
      Model & BLEU$\uparrow$ & Speedup$\uparrow$ \\
      \midrule
      MT & 29.69 & - \\
      Cascade & 22.52 & 1.06$\times$ \\
      \midrule
      ST & 17.44 & 0.87$\times$ \\
      ST + ASR-PT & 20.67 & 1.00$\times$ \\
      \midrule 
      ST + CNN & 20.64 & 1.31$\times$ \\
      \midrule
      ST + Fixed Rate ($k=6$) & 21.14 (83.3\%) & 1.42$\times$ \\
      ST + Fixed Rate ($k=7$) & 20.87 (85.7\%) & 1.43$\times$ \\
      \midrule
      ST + \afst{} & 21.57 (84.4\%) & 1.38$\times$ \\
      ST + \afstf{} & 22.38 (85.1\%) & 1.37$\times$ \\
    \bottomrule
    \end{tabular}
    \caption{BLEU$\uparrow$ and speedup$\uparrow$ on MuST-C En-De test set. $\lambda=0.5$. We evaluate the speedup on GeForce GTX 1080 Ti with a decoding batch size of 16, and report average results over 3 runs. Numbers in parentheses are the sparsity rate. }
    \label{tab:ana_en_de}
\end{table}

\begin{figure}[t]
  \centering
  \small
  \includegraphics[scale=0.40]{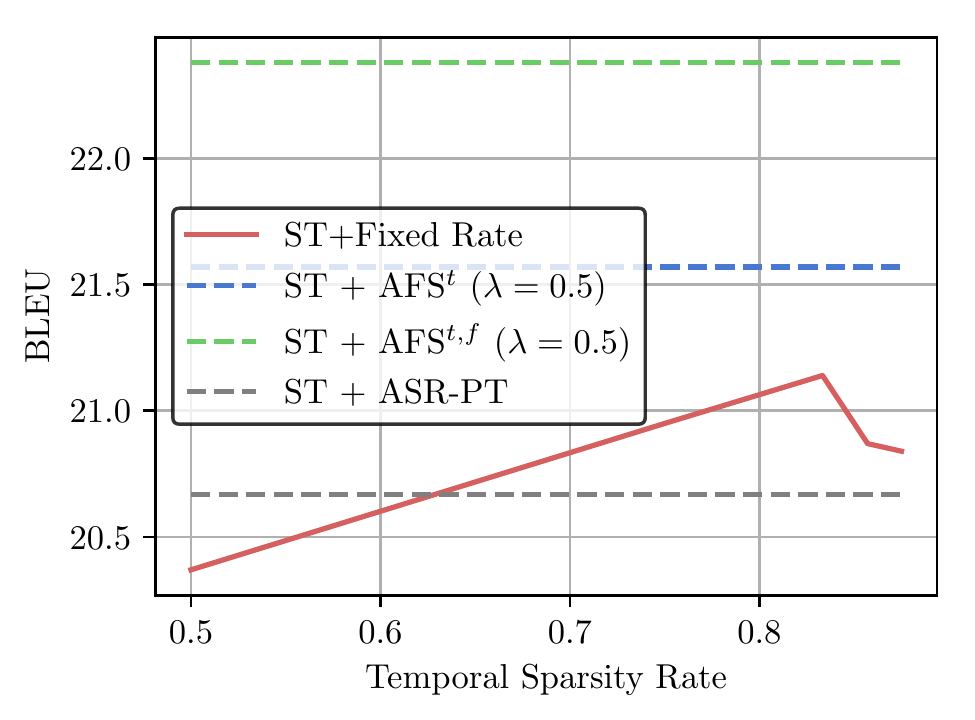}
  \caption{\label{fig:ana_fixed_rate_k} Impact of $k$ in fixed-rate subsampling on ST performance on MuST-C En-De test set. Sparsity rate: $\nicefrac{k-1}{k}$. This subsampling underperforms \afs{}, and degenerates the ST performance at suboptimal rates.}
\end{figure}

We summarize the test results in Table \ref{tab:ana_en_de}, where we set $k=6$ or $k=7$ for \textit{ST+Fixed Rate} with a sparsity rate of around 85\% inspired by our above analysis. Our vanilla ST model yields a BLEU score of 17.44; pretraining on ASR further enhances the performance to 20.67, significantly outperforming the results of \citet{di2019adapting} by 3.37 BLEU. This also suggests the importance of speech encoder pretraining~\cite{di2019adapting,stoian2020analyzing,wang2020curriculum}. We treat ST with ASR-PT as our real baseline. We observe improved translation quality with fixed-rate subsampling, +0.47 BLEU at $k=6$.
Subsampling offers a chance to bypass noisy speech signals and reducing the number of source states makes learning translation alignment easier, but deciding the optimal sampling rate is tough.
Results in Figure~\ref{fig:ana_fixed_rate_k} reveal that fixed-rate subsampling deteriorates ST performance with suboptimal rates. Replacing fixed-rate subsampling with our one-layer CNN also fails to improve over the baseline, although CNN offers more flexibility in feature manipulation. By contrast to fixed-rate subsampling, the proposed \afs{} is data-driven, shifting the decision burden to the data and model themselves. As a result, \afst{} and \afstf{} surpass ASR-PT by 0.9 BLEU and 1.71 BLEU, respectively, substantially narrowing the performance gap compared to the cascade baseline (-0.14 BLEU).

We also observe improved decoding speed: \afs{} runs $\sim$1.37$\times$ faster than ASR-PT. Compared to the fixed-rate subsampling, \afs{} is slightly slower which we ascribe to the overhead introduced by the gating module. Surprisingly, Table \ref{tab:ana_en_de} shows that the vanilla ST runs slower than ASR-PT (0.87$\times$) while the cascade model is slightly faster (1.06$\times$). By digging into the beam search algorithm, we discover that ASR pretraining shortens the number of steps in beam-decoding: $94$ ASR-PT vs. $112$ vanilla ST (on average). The speedup brought by cascading is due to the smaller English vocabulary size compared to the German vocabulary when processing audio inputs.

\begin{figure}[t]
  \centering
  \small
  \includegraphics[scale=0.40]{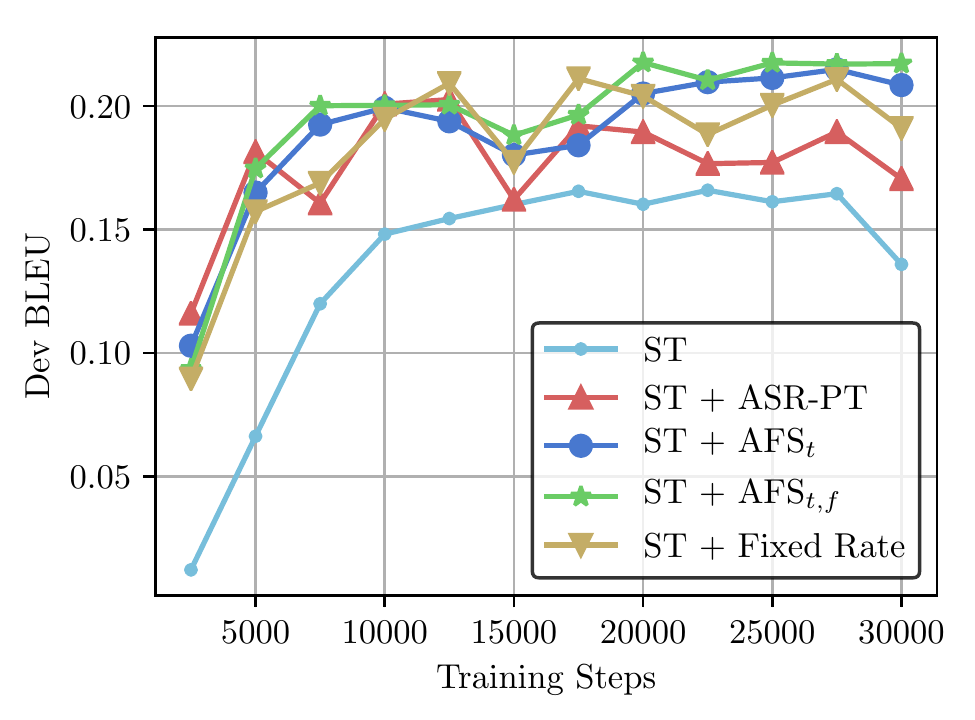}
  \caption{\label{fig:ana_convergence} ST training curves (MuST-C En-De dev set). ASR pretraining significantly accelerates model convergence, and feature selection further stabilizes and improves training. $\lambda=0.5, k=6$.}
\end{figure}

\subsection{Why (Adaptive) Feature Selection?}

Apart from the benefits in translation quality, we go deeper to study other potential impacts of (adaptive) feature selection. We begin with inspecting training curves. Figure \ref{fig:ana_convergence} shows that ASR pretraining improves model convergence; feature selection makes training more stable. Compared to other models, the curve of ST with \afs{} is much smoother, suggesting its better regularization effect.

\begin{figure}[t]
  \centering
  \small
  \includegraphics[scale=0.40]{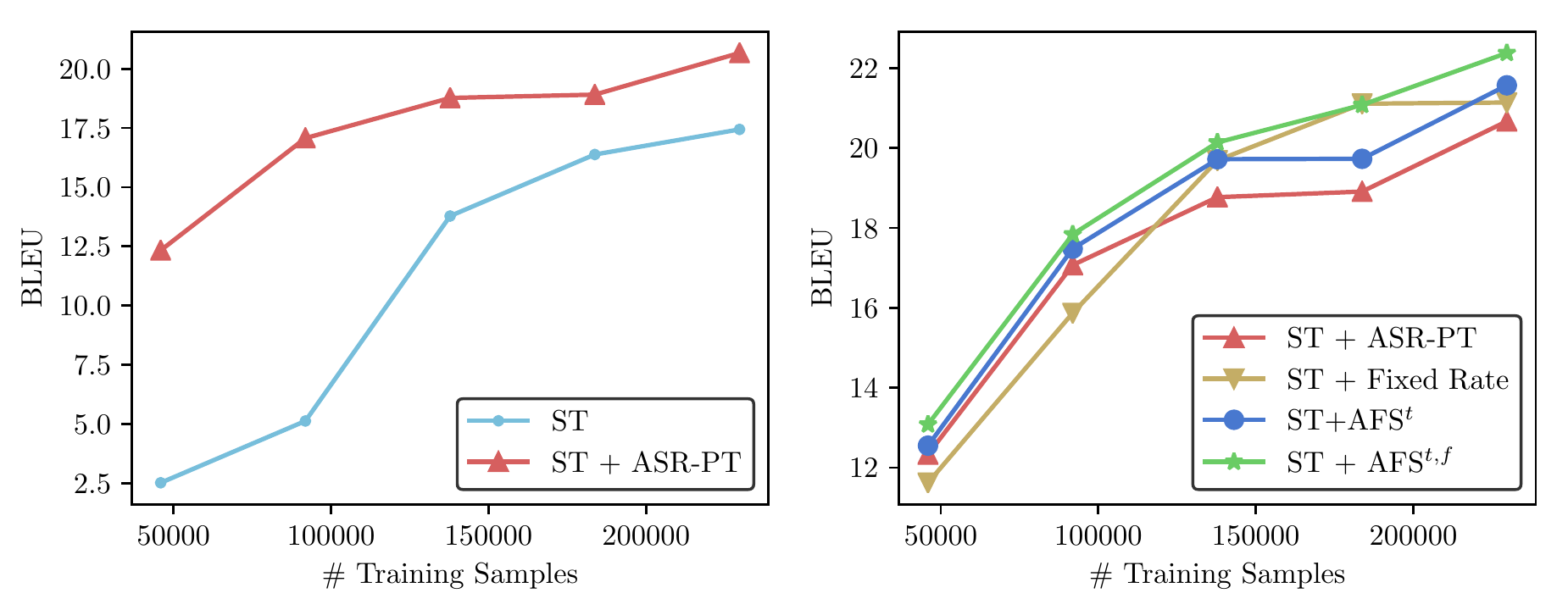}
  \caption{\label{fig:ana_corpus} BLEU as a function of training data size on MuST-C En-De. We split the original training data into non-overlapped five subsets, and train different models with accumulated subsets. Results are reported on the test set. Note that we perform ASR pretraining on the original dataset. $\lambda=0.5, k=6$.}
\end{figure}

We then investigate the effect of training data size, and show the results in Figure \ref{fig:ana_corpus}. Overall, we do not observe higher data efficiency by feature selection on low-resource settings. But instead, our results suggest that feature selection delivers larger performance improvement when more training data is available. With respect to data efficiency, ASR pretraining seems to be more important (Figure \ref{fig:ana_corpus}, left)~\cite{bansal-etal-2019-pre,stoian2020analyzing}. Compared to \afs{}, the fixed-rate subsampling suffers more from small-scale training: it yields worse performance than ASR-PT when data size $\leq 100\text{K}$, highlighting better generalization of \afs{}.

\begin{figure}[t]
  \centering
  \small
  \includegraphics[scale=0.40]{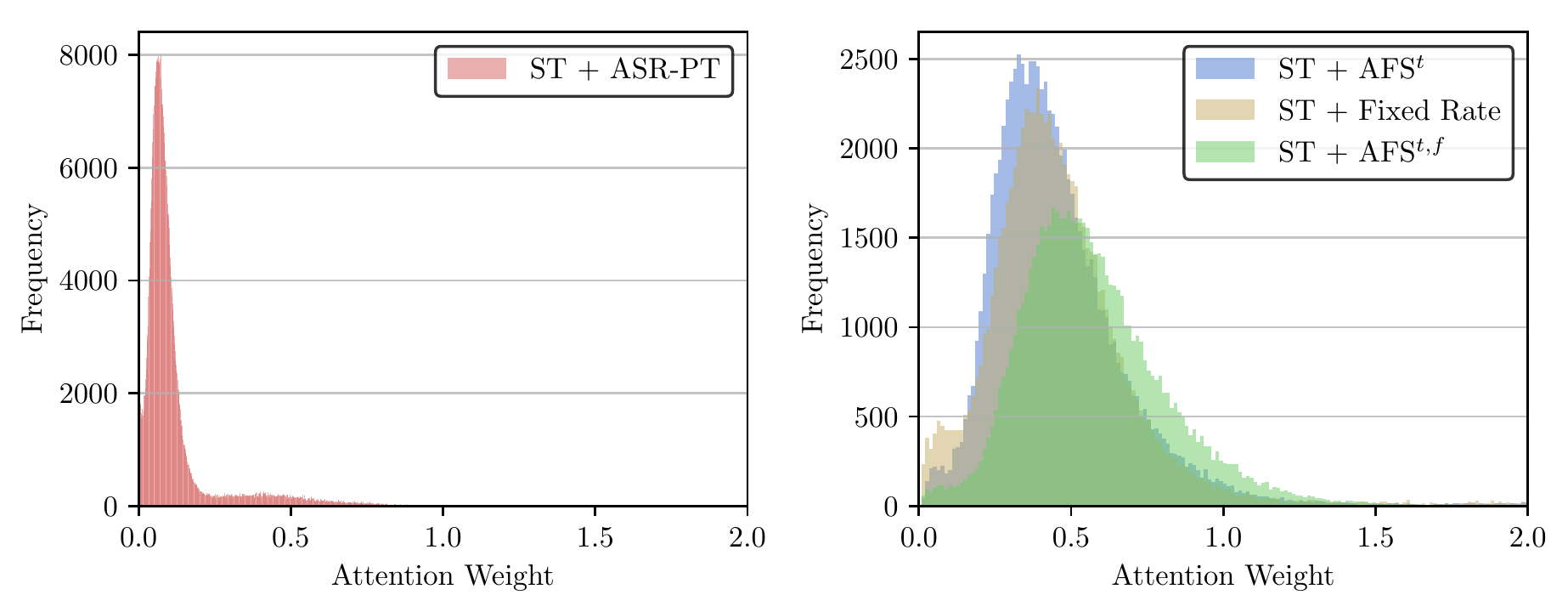}
  \caption{\label{fig:ana_attention} Histogram of the cross-attention weights received per ST encoder output on MuST-C En-De test set. For each instance, we collect attention weights averaged over different heads and decoder layers following~\citet{zhang2020sparsifying}. Larger weight indicates stronger impact of the encoder output on translation. Feature selection biases the distribution towards larger weights. $\lambda=0.5, k=6$.}
\end{figure}

In addition to model performance, we also look into the ST model itself, and focus on the cross-attention weights. Figure \ref{fig:ana_attention} visualize the attention value distribution, where ST models with feature selection noticeably shift the distribution towards larger weights. This suggests that each ST encoder output exerts greater influence on the translation. By removing redundant and noisy speech features, feature selection eases the learning of the ST encoder, and also enhances its connection strength with the ST decoder. This helps bridge the modality gap between speech and text translation.
Although fixed-rate subsampling also delivers a distribution shift similar to \afs{}, its inferior ST performance compared to \afs{} corroborates the better quality of adaptively selected features.

\begin{figure}[t]
  \centering
  \small
    \subcaptionbox{\label{fig:ana_num_feature:duration} Duration Analysis}[0.494\columnwidth]{
        \centering
        \includegraphics[scale=0.40]{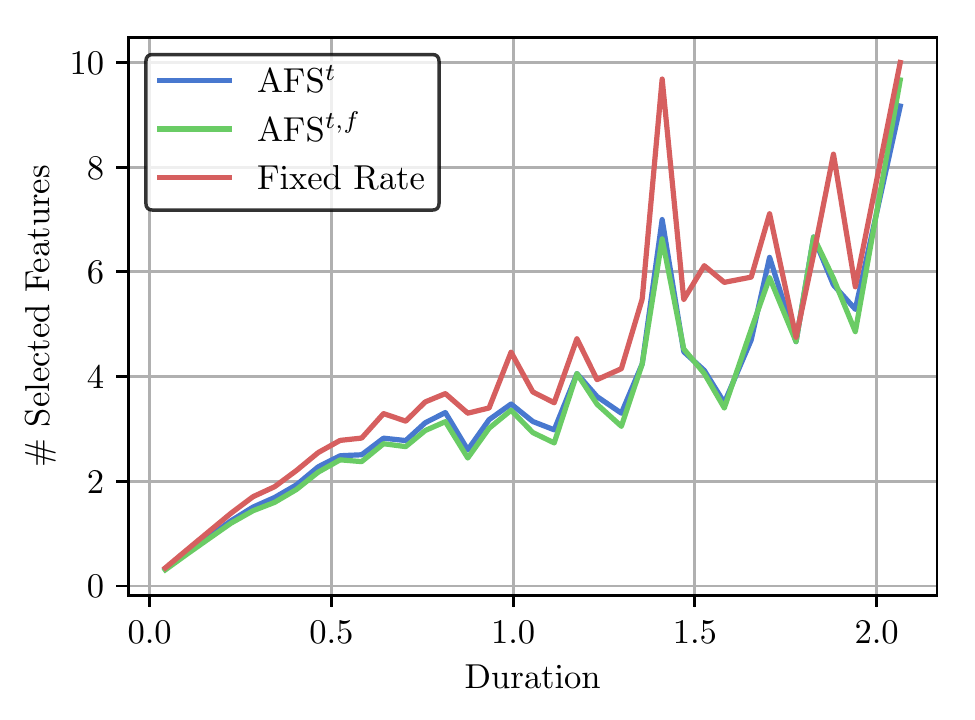}
    }
    \subcaptionbox{\label{fig:ana_num_feature:bin} Position Analysis}[0.494\columnwidth]{
        \centering
        \includegraphics[scale=0.40]{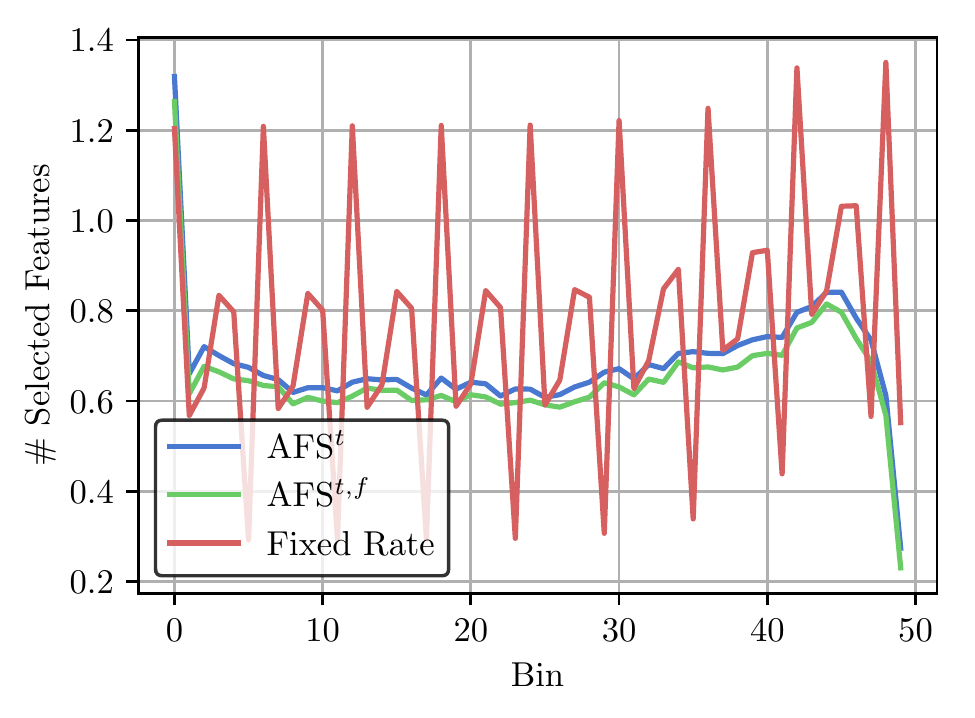}
    }
  \caption{\label{fig:ana_num_feature} The number of selected features vs. word duration (left) and position (right) on MuST-C En-De test set. For word duration, we align the audio and its transcription by Montreal Forced Aligner~\cite{McAuliffe2017}, and collect each words' duration and its corresponding retained feature number. For position, we uniformly split each input into 50 pieces, and count the average number of retained features in each piece. $\lambda=0.5, k=6$.}
\end{figure}

\begin{table*}[t]
    \centering
    \small
    \begin{tabular}{llcccccccc}
      \toprule
      Metric & Model & De & Es & Fr & It & Nl & Pt & Ro & Ru \\
      \midrule
      \multirow{7}{*}{BLEU$\uparrow$}
      & \citet{di2019adapting} & 17.30 & 20.80 & 26.90 & 16.80 & 18.80 & 20.10 & 16.50 & 10.50 \\
      & Transformer + ASR-PT$^*$ & 21.77 & 26.41 & 31.56 & 21.46 & 25.22 & 26.84 & 20.53 & 14.31 \\
      \cmidrule{2-10}
      & ST & 17.44 & 23.85 & 28.43 & 19.54 & 21.23 & 22.55 & 17.66 & 12.10 \\
      & ST + ASR-PT & 20.67 & 25.96 & 32.24 & 20.84 & 23.27 & 24.83 & 19.94 & 13.96  \\
      & Cascade & 22.52 & 27.92 & 34.53 & 24.02 & 26.74 & 27.57 & 22.61 & 16.13 \\
      \cmidrule{2-10}
      & ST + \afst{} & 21.57 & 26.78 & 33.34 & 23.08 & 24.68 & 26.13 & 21.73 & 15.10 \\
      & ST + \afstf{} & 22.38 & 27.04 & 33.43 & 23.35 & 25.05& 26.55 & 21.87 &  14.92 \\
      \midrule
      \multirow{2}{*}{SacreBLEU $\uparrow$} 
      & ST + \afst{} & 21.6\phantom{0} & 26.6\phantom{0} & 31.5\phantom{0} & 22.6\phantom{0} & 24.6\phantom{0} & 25.9\phantom{0} & 20.8\phantom{0} & 14.9\phantom{0} \\
      & ST + \afstf{} & 22.4\phantom{0} & 26.9\phantom{0} & 31.6\phantom{0} & 23.0\phantom{0} & 24.9\phantom{0} & 26.3\phantom{0} & 21.0\phantom{0} &  14.7\phantom{0} \\
      \midrule
      Temporal & ST + \afst{} & 84.4\% & 84.5\% & 83.2\% & 84.9\% & 84.4\% & 84.4\% & 84.7\% & 84.2\% \\
      Sparsity Rate & ST + \afstf{} &  85.1\% & 84.5\% & 84.7\% & 84.9\% & 83.5\% & 85.1\% & 84.8\% & 84.7\% \\
      \midrule
      \multirow{2}{*}{Speedup $\uparrow$}
      & ST + \afst{} & 1.38$\times$ & 1.35$\times$ & 1.50$\times$ & 1.34$\times$ & 1.54$\times$ & 1.43$\times$ & 1.59$\times$ & 1.31$\times$\\
      & ST + \afstf{} & 1.37$\times$ & 1.34$\times$ & 1.50$\times$ & 1.39$\times$ & 1.42$\times$ & 1.26$\times$ & 1.46$\times$ & 1.37$\times$ \\
    \bottomrule
    \end{tabular}
    \caption{Performance over 8 languages on MuST-C dataset. $^*$: results reported by the ESPNet toolkit~\cite{watanabe2018espnet}, where the hyperparameters of beam search are tuned for each dataset.}
    \label{tab:res_must_c}
\end{table*}

\paragraph{\afs{} vs. Fixed Rate} We compare these two approaches by analyzing the number of retained features with respect to word duration and temporal position. Results in Figure \ref{fig:ana_num_feature:duration} show that the underlying pattern behind these two methods is similar: words with longer duration correspond to more speech features. However, when it comes to temporal position, Figure \ref{fig:ana_num_feature:bin} illustrates their difference: fixed-rate subsampling is context-independent, periodically picking up features; while \afs{} decides feature selection based on context information. The curve of \afs{} is more smooth, indicating that features kept by \afs{} are more uniformly distributed across different positions, ensuring the features' informativeness.

\begin{figure}[t]
  \centering
  \small
  \includegraphics[scale=0.40]{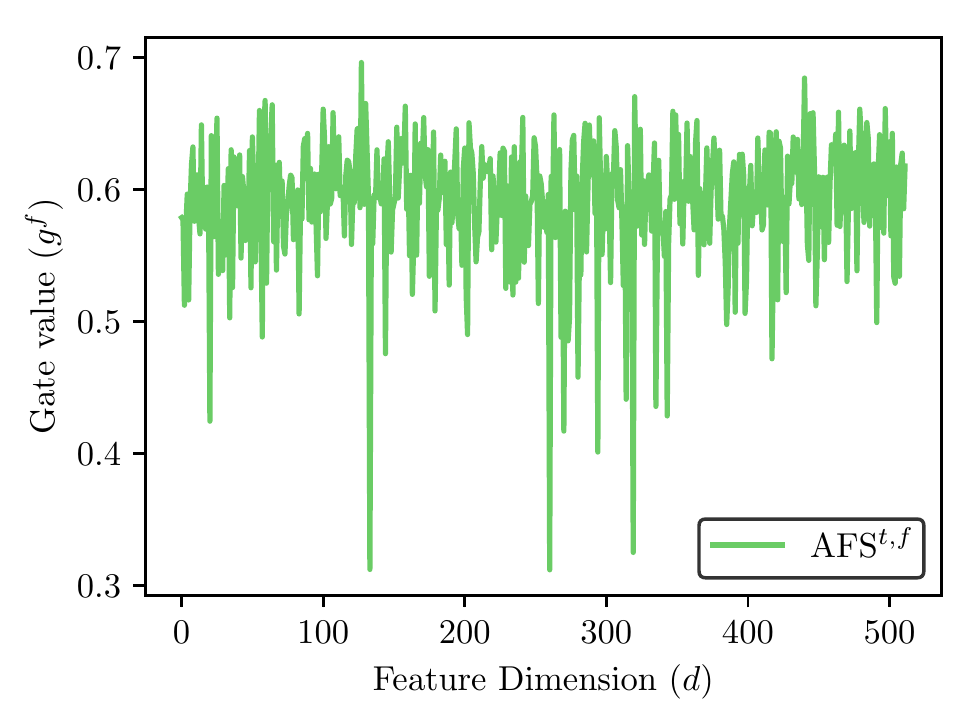}
  \caption{\label{fig:ana_gate_f} Illustration of feature gate $\mathbf{g}^f$ with $\lambda=0.5$.}
\end{figure}
\paragraph{\afst{} vs. \afstf{}} Their only difference lies at the feature gate $\mathbf{g}^f$. We visualize this gate in Figure \ref{fig:ana_gate_f}. Although this gate induces no sparsification, it offers \afstf{} the capability of adjusting the weight of each neuron. In other words, \afstf{} has more freedom in manipulating speech features.

\begin{table}[t]
    \centering
    \small
    \begin{tabular}{llc}
      \toprule
      Metric & {Model} & En-Fr \\
      \midrule
      \multirow{9}{*}{BLEU$\uparrow$} 
      & {\citet{berard2018end}} & 13.40 \\
      & {\citet{watanabe2018espnet}} & 16.68  \\
      & {\citet{liu2019end}} & 17.02  \\
      & {\citet{wang2019bridging}} & 17.05 \\
      & {\citet{wang2020curriculum}} & 17.66 \\
      \cmidrule{2-3}
      & {ST} & 14.32  \\
      & {ST + ASR-PT} & 17.05  \\
      & {Cascade} & 18.27  \\
      \cmidrule{2-3}
      & ST + \afst{} & 18.33 \\
      & ST + \afstf{} & 18.56 \\
      \midrule
      \multirow{2}{*}{SacreBLEU$\uparrow$}  
      & ST + \afst{} & 16.9\phantom{0} \\
      & ST + \afstf{} & 17.2\phantom{0} \\
      \midrule
      Temporal
      & ST + \afst{} & 84.7\% \\
       Sparsity Rate
      & ST + \afstf{} & 83.5\% \\
      \midrule
      \multirow{2}{*}{Speedup$\uparrow$} 
      & ST + \afst{} & 1.84$\times$ \\
      & ST + \afstf{} & 1.78$\times$ \\
    \bottomrule
    \end{tabular}
    \caption{Performance on LibriSpeech En-Fr.}
    \label{tab:res_libri_trans}
\end{table}

\subsection{Results on MuST-C and LibriSpeech}

Table \ref{tab:res_must_c} and Table \ref{tab:res_libri_trans} list the results on MuST-C and LibriSpeech En-Fr, respectively. Over all tasks, \afst{}/\afstf{} substantially outperforms ASR-PT by 1.34/1.60 average BLEU, pruning out 84.5\% temporal speech features on average and yielding an average decoding speedup of 1.45$\times$. Our model narrows the gap against the cascade model to -0.8 average BLEU, where \afs{} surpasses Cascade on LibriSpeech En-Fr, without using KD~\cite{liu2019end} and data augmentation~\cite{wang2020curriculum}.
Comparability to previous work is limited due to possible differences in tokenization and letter case.
To ease future cross-paper comparison, we provide SacreBLEU~\cite{post-2018-call}\footnote{signature: BLEU+c.mixed+\#.1+s.exp+tok.13a+version.1.3.6} for our models.

\section{Related Work}

\paragraph{Speech Translation} Pioneering studies on ST used a cascade of separately trained ASR and MT systems~\cite{ney1999speech}. Despite its simplicity, this approach inevitably suffers from mistakes made by ASR models, and is error prone. Research in this direction often focuses on strategies capable of mitigating the mismatch between ASR output and MT input, such as representing ASR outputs with lattices~\cite{saleem2004using,mathias2006statistical,zhang-etal-2019-lattice,beck-etal-2019-neural}, injecting synthetic ASR errors for robust MT~\cite{tsvetkov-etal-2014-augmenting,cheng-etal-2018-towards} and differentiable cascade modeling~\cite{kano2017structured,anastasopoulos-chiang-2018-tied,sperber-etal-2019-attention}.

In contrast to cascading, another option is to perform direct speech-to-text translation. \citet{duong-etal-2016-attentional} and \citet{berard2016listen} employ the attentional encoder-decoder model~\cite{DBLP:journals/corr/BahdanauCB14} for E2E ST without accessing any intermediate transcriptions. E2E ST opens the way to bridging the modality gap directly, but it is data-hungry, sample-inefficient and often underperforms cascade models especially in low-resource settings~\cite{bansal2018low}. This led researchers to explore solutions ranging from efficient neural architecture design~\cite{karita2019transformerasr,di2019adapting,sung2019towards} to extra training signal incorporation, including multi-task learning~\cite{weiss2017sequence,liu2019synchronous}, submodule pretraining~\cite{bansal-etal-2019-pre,stoian2020analyzing,wang2020curriculum}, knowledge distillation~\cite{liu2019end}, meta-learning~\cite{indurthi2019data} and data augmentation~\cite{kocabiyikoglu-etal-2018-augmenting,jia2019leveraging,pino2019harnessing}. Our work 
focuses on E2E ST, but we investigate feature selection which has rarely been studied before.

\paragraph{Speech Feature Selection} Encoding speech signals is challenging as acoustic input is lengthy, noisy and redundant. To ease model learning, previous work often selected features via downsampling techniques, such as convolutional modeling~\cite{di2019adapting} and fixed-rate subsampling~\cite{lu2015study}. Recently, \citet{Zhang2019trainable} and \citet{na2019adaptive} proposed dynamic subsampling for ASR which learns to skip uninformative features during recurrent encoding. Unfortunately, their methods are deeply embedded into recurrent networks, hard to adapt to other architectures like Transformer~\cite{NIPS2017_7181_attention}. 
Recently, \citet{salesky-etal-2019-exploring} have explored phoneme-level representations for E2E ST, which reduces speech features temporarily by $\sim$80\% and obtains significant performance improvement, but this requires non-trivial phoneme recognition and alignment.

Instead, we resort to sparsification techniques which have achieved great success in NLP tasks recently~\cite{correia-etal-2019-adaptively,child2019generating,zhang2020sparsifying}. In particular, we employ \lzerodrop{}~\cite{zhang2020sparsifying} for \afs{} to dynamically retain informative speech features, which is fully differentiable and independent of concrete encoder/decoder architectures. We extend \lzerodrop{} by handling both temporal and feature dimensions with different gating networks, and apply it to E2E ST.

\section{Conclusion and Future Work}

In this paper, we propose adaptive feature selection for E2E ST to handle redundant and noisy speech signals. We insert \afs{} in-between the ST encoder and a pretrained, frozen ASR encoder to filter out uninformative features contributing little to ASR. We base \afs{} on \lzerodrop{}~\cite{zhang2020sparsifying}, and extend it to modeling both temporal and feature dimensions. Results show that \afs{} improves translation quality and accelerates decoding by $\sim$1.4$\times$ with an average temporal sparsity rate of $\sim$84\%. \afs{} successfully narrows or even closes the performance gap compared to cascading models. 

While most previous work on sparsity in NLP demonstrates its benefits from efficiency and/or interpretability perspectives~\cite{zhang2020sparsifying}, we show that sparsification in our scenario -- E2E ST -- leads to substantial performance gains.

In the future, we will work on adapting \afs{} to simultaneous speech translation.

\section*{Acknowledgments}

We would like to thank Shucong Zhang for his great support on building our ASR baselines. IT acknowledges
support of the European Research Council (ERC Starting grant 678254) and the Dutch National Science Foundation
(NWO VIDI 639.022.518).
This work has received funding from the European Union’s Horizon 2020 Research and
Innovation Programme under Grant Agreement No 825460 (ELITR).
Rico Sennrich acknowledges support of the Swiss National Science Foundation (MUTAMUR; no.\ 176727).

\bibliographystyle{acl_natbib}
\bibliography{emnlp2020}

\end{document}